\documentclass[conference]{IEEEtran}
\IEEEoverridecommandlockouts
\usepackage{cite}
\usepackage{amsmath,amssymb,amsfonts}
\usepackage{algorithmic}
\usepackage{graphicx}
\usepackage{textcomp}
\usepackage{multirow}
\usepackage[table,xcdraw]{xcolor}
\def\BibTeX{{\rm B\kern-.05em{\sc i\kern-.025em b}\kern-.08em
    T\kern-.1667em\lower.7ex\hbox{E}\kern-.125emX}}
\begin{document}

\title{Reinforcement Learning Agent Design and Optimization with Bandwidth Allocation Model\\
\thanks{Research supported by ANIMA Institute.}
}

\author{\IEEEauthorblockN{Rafael F. Reale}
\IEEEauthorblockA{ \textit{Federal Institute of Bahia (IFBA - Valença)}, Brazil \\
https://orcid.org/0000-0003-4464-4548}
\and
\IEEEauthorblockN{Joberto S. B. Martins}
\IEEEauthorblockA{\textit{Salvador University (UNIFACS)}, Brazil \\
https://orcid.org/0000-0003-1310-9366}
\and

}

\maketitle

\begin{abstract}

Reinforcement learning (RL) is currently used in various real-life applications. RL-based solutions have the potential to generically address problems, including the ones that are difficult to solve with heuristics and meta-heuristics and, in addition, the set of problems and issues where some intelligent or cognitive approach is required. However, reinforcement learning agents require a not straightforward design and have important design issues. RL agent design issues include the target problem modeling, state-space explosion, the training process, and agent efficiency. Research currently addresses these issues aiming to foster RL dissemination. A BAM model, in summary, allocates and shares resources with users. There are three basic BAM models and several hybrids that differ in how they allocate and share resources among users. This paper addresses the issue of an RL agent design and efficiency. The RL agent's objective is to allocate and share resources among users. The paper investigates how a BAM model can contribute to the RL agent design and efficiency. The AllocTC-Sharing (ATCS) model is analytically described and simulated to evaluate how it mimics the RL agent operation and how the ATCS can offload computational tasks from the RL agent. The essential argument researched is whether algorithms integrated with the RL agent design and operation have the potential to facilitate agent design and optimize its execution. The ATCS analytical model and simulation presented demonstrate that a BAM model offloads agent tasks and assists the agent's design and optimization.

\end{abstract}


\section{Introduction}\label{sec:Introduction}

Reinforcement learning \cite{sutton_introduction_1998} is being extensively adopted for the solution and optimization of problems for real-life applications like 5G and beyond (5G\&B), the internet of things (IoT), internet of vehicles (IoV), and network slicing \cite{wu_ai-native_2022} \cite{kumar_iot_2022} \cite{afrin_resource_2021} \cite{ji_survey_2020}.

The fostering of machine learning use in various areas depends, at least in part, on having agent modeling and design approaches or methods that make ML agents competitive concerning heuristics and meta-heuristics. In effect, it is expected that ML agents with formal, framework-supported, or speedy-like approaches could not only solve the unsolved problem using heuristics but also address the current problems that require intelligence or knowledge acquisition.

The design of machine learning agents is a fundamental issue currently addressed by research. As such, the design of machine learning agents needs frameworks, methods, or approaches to facilitate and optimize agent design.

Bandwidth allocation models (BAM), in summary, share resources among users based on resource priority \cite{martins_uma_2015}. The BAM models have attributes that allow controlling the allocation and sharing of resources with different characteristics (behaviors).

This paper addresses the issue of using a bandwidth allocation model (BAM) as part of the reinforcement learning agent deployment. It investigates how the BAM model can support an RL agent design and efficiency. The BAM capability to offload computational tasks from the RL agent is evaluated with the objective to simplify the agent modeling task and allow the agent operation optimization aiming to facilitate and improve agent design and operation.

The remainder of the paper is structured as follows. Section \ref{sec:RelatedWork} presents the related work and Sections \ref{sec:RLAgentModel} and \ref{sec:BAMSummary} are summary background sections about the RL agent design issues and bandwidth allocation models operation and management. Sections \ref{sec:BAMAnalyticalModel} presents an analytical model for the BAM ATCS. Section \ref{sec:ATCSUseCase} illustrates the ATCS operation dynamics with a use case simulation and Section \ref{sec:RLandBAMSupport} discusses the ATCS support for RL agent design and optimization. Finally, section \ref{sec:FinalConsiderations} concludes with an overview of the main highlights, contributions, and future work.

\begin{figure*}[htbp]
    \centering
    \includegraphics[width=0.5\textwidth]{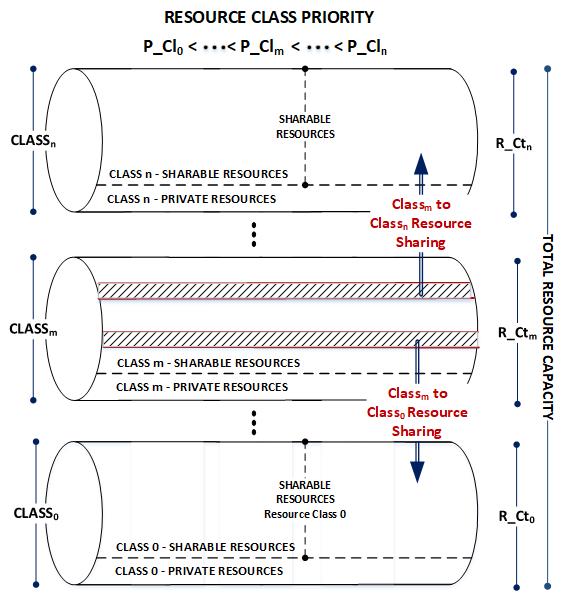}
    \caption{BAM Resource Allocation and Resource Sharing.}
    \label{fig:BAM_Sharing}
\end{figure*}

\section{Related Work}\label{sec:RelatedWork}

Reinforcement learning agents are being developed for various application areas and the issues and problems related to these developments are mentioned in the literature and are the focus of attention \cite{ottoni_tuning_2020}.

The state-space explosion issue impacts substantially the RL agent and there are approaches that are used to mitigate this problem that limits agent state-space \cite{wong_state-space_2021}, or use auxiliary processing to do so \cite{zhao_fuzzy_2010} \cite{jiang_improved_2019}. Most of the time, these approaches are presented as part of the agent definition and deployment not having specific regard or more extensive discussions about it.

The agent design and optimization approach proposed in this paper is aligned with the strategy of using auxiliary algorithms, methods, or software to reduce the working state-space of the agent \cite{ha_reinforcement_2019} \cite{nguyen_multi-objective_2020}.

The proposal uses the bandwidth allocation models (BAMs) to mitigate the state-space explosion issue in an RL agent and, to the limit of our knowledge, this specific approach is new in the literature.

Machine learning agents in general are adopting new development and deployment approaches like distributing agents (multi-agent RL) \cite{zhang_multi-agent_2021} \cite{iqbal_actor-attention-critic_2019} and sharing knowledge among agents (federated learning) \cite{zhang_survey_2021} \cite{lim_federated_2020}. The pushing of machine learning agents' design into new development strategies is, to some extent, motivated by the need to enforce efficiency.

\section{Reinforcement Learning Agent Modeling and Training}\label{sec:RLAgentModel}

Reinforcement learning agents are being used extensively for a broad range of problems with multiple input variables and multiple objectives. In such scenarios, heuristics or meta-heuristics are difficult to develop and apply and, consequently, knowledge acquisition with machine learning becomes of paramount importance.

However, machine learning agents need to be modeled to the target problem aiming to solve it. Like any heuristic or meta-heuristic algorithm design, the agent modeling task or agent design requires minimal expertise concerning the addressed problem and a set of development and experimentation steps.

There are frameworks and methods for agent design like the design of experiments (DoE) methodology \cite{montgomery_design_2017}, the ANOVA statistical method for data analysis \cite{stahle_analysis_1989}, and, obviously, the ad-hoc method in which a trial-and-error design is used. 

The steps involved in the reinforcement learning agent design for a target problem are as follows \cite{montgomery_design_2017}:
\begin{itemize}
    \item Definition and model of the relevant agent input variables (factors) and  output variables (variables of interest);
    \item Experiment design and test runs; and
    \item Statistical analysis of data.
\end{itemize}

Some of the reinforcement problems and issues concerning agent modeling, training, and efficiency include:

\begin{itemize}
    \item The agent designing process;
    \item The agent state-space explosion; and
    \item RL parameters definition like the reward matrix and RL learning parameters tuning.
\end{itemize}

Agent design, state-space explosion control, and agent parameters definition and tuning are large and complex problems. This paper aims to contribute to facilitating the design process and reducing the state-space explosion for an RL agent.

\subsection{The Proposed Approach to Facilitate Agent's Design and Contribute to Agent´s Efficiency}

The essential aspect investigated is the inclusion of a BAM model software in an RL agent deployment to evaluate to what extent the BAM model can offload computational tasks that are, primarily, executed by the RL agent software.

In following this approach, it is necessary to understand the essential aspects of BAM model operation and to what extent it can mimic the behavior and operation of an RL agent.

\section{Bandwidth Allocation Model Summary}\label{sec:BAMSummary}

The bandwidth allocation models basically allocate and share resources among users considering their priority \cite{martins_uma_2015}. There are three basic BAM models and several hybrids that basically differ in how they allocate and share resources among users.

The three basic bandwidth allocation models that are considered in this discussion are:
\begin{itemize}
    \item The Maximum Allocation Model (MAM)\cite{faucher_maximum_2005};
    \item The Russian Dolls Model (RDM) \cite{da_costa_pinto_neto_adapt-rdm_2008}; and
    \item The AllocTCSharing (ATCS) model \cite{reale_alloctc-sharing_2011}.
\end{itemize}

All BAM models group resources in resource classes with priorities and each of the BAM mentioned above has resource allocation and sharing characteristics as follows:

\begin{itemize}
    \item In the MAM model, resources are grouped into isolated classes that never share their resources with other resource classes;
    \item In the RDM model, resources belonging to high-priority classes can be shared with lower-priority resource classes, and
    \item In the AllocTCSharing (ATCS) model, resources are shared among all resource classes regardless of their priority.
\end{itemize}

Figure \ref{fig:BAM_Sharing} illustrates the BAM resource allocation and sharing operations commonly used by the different BAM models.

It is essential to highlight that BAM resource allocation is meaningful only when there are limited resource constraints. In other words, some resource sharing is required to comply with users' demands dynamically.

\subsection{BAM Parameters for Resource Allocation}

\begin{figure}[htbp]
\centerline{\includegraphics[width=0.5\textwidth]{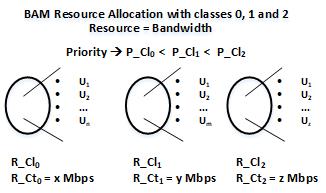}}
\caption{BAM resource class ($R\_Cl_y$) parameters setup: resource constraint ($R\_Ct_x$) and priority ($P\_Cl_j$) by class.}
\label{fig:BAMParameters}
\end{figure}

The bandwidth allocation model parameters used to model the target problem and to manage BAM operation are as follows (Figure \ref{fig:BAMParameters}):

\begin{itemize}
    \item Class resource constraint ($R\_Ct_y$); and
    \item Class priorities  ($P\_Cl_j$);
\end{itemize}

The class resource constraint is the number or amount of resources available to the class. Users request resources from classes.

Although the application of BAM models has focused initially on sharing the resource bandwidth (link bandwidth sharing problem), it is worth highlighting that BAM resources may be generalized to allocate and share various discrete units used by computational systems like slots for elastic optical networks (EON) \cite{duraes_evaluating_2017} and slice resource management \cite{el-mekkawi_squatting_2020}, among others.

\section{The ATCS Analytical Model}\label{sec:BAMAnalyticalModel}

\begin{figure*}[htbp]
    \centering
    \includegraphics[width=0.7\textwidth]{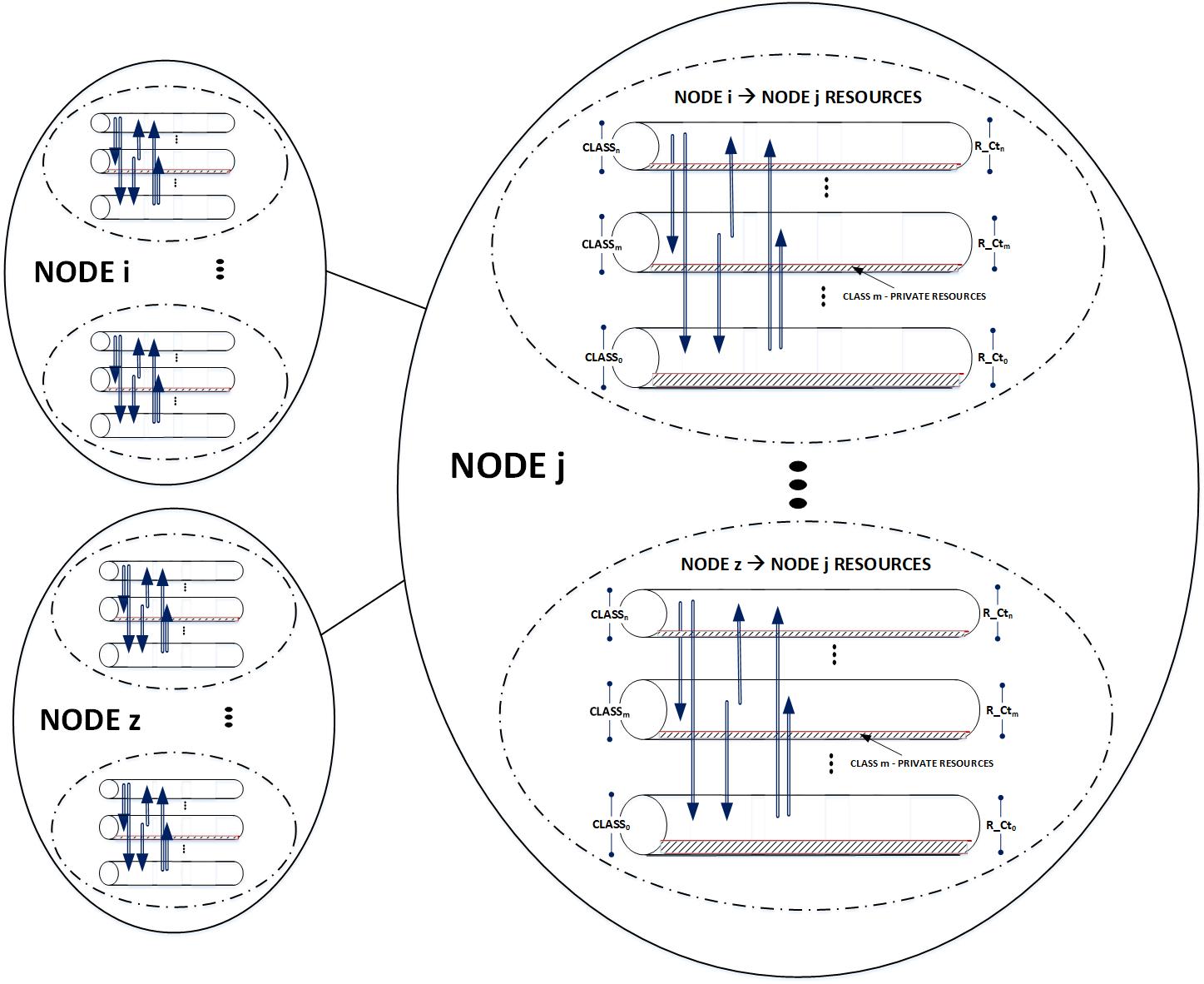}
    \caption{ATCS Resource Allocation and Resource Sharing among Resource Classes.}
    \label{fig:ATCS_Sharing}
\end{figure*}

The BAM models (MAM, RDM, and ATCS) essentially share resources among clients based on their priority. A formal model describing their operation is necessary to detail how they allocate and share resources. The BAM model focused on in this paper is the ATCS since it has a flexible approach to allocating and sharing resources and, as such, it potentially shows a closer behavior to an RL agent.

The AllocTC-Sharing (ATCS) analytical model presents the resource allocation and sharing behavior executed by the model. It aims to identify to what extent the ATCS mimic the behavior of a reinforcement learning agent allocating and sharing resources among user with similar objectives as defined for the ATCS model. The ATCS model is the target of the analytical presentation because it provides the best possible method to allocate and share resources among clients as discussed in \cite{oliveira_cognitive_2018}. The ATCS choice is also motivated by the fact that ATCS has more capabilities in terms of resource sharing and consequently may potentially be closer to an RL agent's behavior.


The network infrastructure for ATCS is modeled as a bidirectional graph ${\aleph} ^q= \left( N^q, L^q\right)$, where $N^q$ = $\{n_1, n_2, n_3, ..., n_n\}$ is the set of nodes in the network $q$ and $L^q$ = $\{l_{ij}\}$, with $i$ and $j$ = $1, 2,3, ..., n$, is the set of connections or links, with link $l_{ij}$ connecting $n_i$ to $n_j$.

The node connectivity matrix (NCM) is defined by $C=[c_{ij}]$ with $i$ and $j$ $=1,2,3,..., n$, and  $c_{ij}=1$ if $n_i$ is connected to $n_j$; $0$ otherwise and $c_{ii}=0$ $\forall i$ by definition.


The node resource matrix (NRM) for the bidirectional network is defined by $NR=[NR_{ij}]$ with $i$ and $j$ $=1,2,3,..., n$, where $NR_{ij}$ is the total amount of resources of  node $n_{ij}$ managed by the ATCS to allocate all incoming resource requests arriving at $n_i$ to $n_j$.

This modeling approach allows a generalization for the type of resources in a node that is available for another connected node for both directions in terms of the node connectivity ($n_{ij}$ and  $n_{ji}$).



The ATCS defines for each connected node $n_{ij}$ (node $n_{i}$ connected to $n_{j}$) a set of resource classes $(R\_CL^k)$, with $k$ $=0,1,..., n$. Each resource class $R\_CL^k_{ij}$ belonging to the node $n_{ij}$ has a resource constraint $R\_CT^k_{ij}$ (Resource Class Constraint), such that:
\begin{equation}
NRC_{ij}\ge\sum_{i=1}^{n}\sum_{j=1}^{n}\sum_{k=1}^{n} R\_CT^k_{ij}
\end{equation}

Where:
\begin{itemize}
    \item $NRC_{i}$ is the total node resource capacity available at $n_j$ to be allocated to connected node $n_i$, and
    \item Each connected node has a set of classes with a limited amount of resources ($R\_CT^k_{ij}$). 
\end{itemize}



Each node can consume resources from the next node in the path (bidirectional path). As such, resources might be allocated in a sequence of nodes like, for instance, packets (traffic) flowing through a series of nodes allocating resources (bandwidth) in the path. Again, it is essential to highlight that with this generalization approach, resources, independently of their type, can be allocated at network nodes.

In the ATCS operation by node, $R_{ij}R\_Cl^kN^j$ is the incoming resource request from node $i$, with $i=1,2,..,n$, belonging to resource class $R\_CL^k$, with $k=0,1,...,n$ at node $n_{j}$.



Resources are dynamically allocated by class and the ATCS defines the resource class ($R\_Cl$) used or allocated resources ($R\_Cl\_UR$). The  $R\_Cl\_UR$  matrix is defined by $R\_Cl\_UR$$=[R\_Cl^{k}{UR_{ij}}]$, with $k=0,1,...,n$, and $i$ and $j$ $=1,2,..,n$.

The bandwidth constraint for the resource classes is:
\begin{equation}
R\_Cl^kUR_{ij}\le R\_CT^k_{ij} \forall k,i,j
\end{equation}

\begin{itemize}
    \item Resource classes ($R\_Cl^s$) have a priority;
    \item Resource sharing is allowed between all resource classes; and
    \item The resources shared are limited by the amount of resources defined for the resource class ($R\_CT$).
\end{itemize}

In addition to the resource class capacity and allocation control, the ATCS model configuration and operation includes a resource class priority for all links $l_{ij}$ with $i,j = 1,2, ..., n$ as follows:

\begin{equation} \label{eq:pri}
R\_CL_p^0 > R\_CL_p^1 > R\_CL_p^2 > \dots > R\_CL_p^n
\end{equation}

In the ATCS convention, resource class $0$ priority ($R\_CL_p^0$) has the highest value, and resource class $n$ priority ($R\_CL_p^n$) has the lowest value.

The resource class priority is used by the ATCS to:

\begin{itemize}
     \item Allocate unused resources from other resource classes; and
     \item Preempt or return shared resources previously allocated from other resource classes.
  
\end{itemize}

ATCS policy allows each resource class $R\_CL^k$ to have two resource partitions: private and public. The private resource partition or pool is, by definition, exclusively allocated for $R\_CL^k$ users, and the public resources may be allocated by other resource classes when not used by  $R\_CL^k$ users. As such for all $l_{ij}$ with $i,j = 1,2, ..., n$:
\begin{equation}
R\_Cl^kR_{ij}= RCl^k_{pri}R_{ij}+RCl^k_{pub}R_{ij}
\end{equation}

Where:
\begin{itemize}
    \item $R\_Cl^kR_{ij}$ is the amount of resources available for $R\_Cl^k$ at node $n_j$ for node $n_i$ 
\end{itemize}

\begin{table}[]
 \caption{ATCS dynamics allocating and sharing resources}
 \label{Tab:LSPsEstabelecidas}
  \centering
\begin{tabular}{|l|l|l|l|l|}
\hline
\rowcolor[HTML]{C0C0C0} 
                      & Constraint & Request  & Resource Class & Resource                                   \\ \hline
&   & \cellcolor[HTML]{34FF34}Request 3 & \cellcolor[HTML]{34FF34}$R\_Cl^2$ & \cellcolor[HTML]{34FF34}10 Mbps \\ 

\cline{3-5} 
\multirow{-2}{*}{$R\_CT^2_{ij}$} & \multirow{-2}{*}{20 Mbps} & \cellcolor[HTML]{FFCC67}                        & \cellcolor[HTML]{FFCC67}                       & \cellcolor[HTML]{FFCC67}                          \\ \cline{1-2}
                      &                           & \multirow{-2}{*}{\cellcolor[HTML]{FFCC67}Request 5} & \multirow{-2}{*}{\cellcolor[HTML]{FFCC67}$R\_Cl^2$} & \multirow{-2}{*}{\cellcolor[HTML]{FFCC67}20 Mbps} \\ \cline{3-5} 
                      &                           & \cellcolor[HTML]{34FF34}Request 2                   & \cellcolor[HTML]{34FF34}$R\_Cl^1$                    & \cellcolor[HTML]{34FF34}30 Mbps                   \\ \cline{3-5} 
\multirow{-3}{*}{$R\_CT^1_{ij}$} & \multirow{-3}{*}{50 Mbps} & \cellcolor[HTML]{FFCC67}                        & \cellcolor[HTML]{FFCC67}                       & \cellcolor[HTML]{FFCC67}                          \\ \cline{1-2}
                      &                           & \multirow{-2}{*}{\cellcolor[HTML]{FFCC67}Request 4} & \multirow{-2}{*}{\cellcolor[HTML]{FFCC67}$R\_Cl^0$}  & \multirow{-2}{*}{\cellcolor[HTML]{FFCC67}30 Mbps} \\ \cline{3-5} 
\multirow{-2}{*}{$R\_CT^0_{ij}$} & \multirow{-2}{*}{30 Mbps} & \cellcolor[HTML]{34FF34}Request 1                   & \cellcolor[HTML]{34FF34}$R\_Cl^0$                    & \cellcolor[HTML]{34FF34}20 Mbps                   \\ \hline
\end{tabular}
\end{table}

In case private and public resource pools are considered in the ATCS operation, the resource class used resources ($R\_Cl\_UR$) matrix must then be rewritten as follows for all $l_{ij}$ with $i,j = 1,2, ..., n$:
\begin{equation} \label{eq:ATCS_Operation}
\begin{split}
R\_Cl^kUR_{ij}= [R\_Cl^k_{pri}UR_{ij}+R\_Cl^k_{pub}UR_{ij} \\
+\sum_{z=1}^{n}R\_Cl^k_{pub}UR_{ij}]
\forall z\neq k
\end{split}
\end{equation}

Equation \ref{eq:ATCS_Operation} describes the ATCS operation when allocating and sharing resources among resource classes. In summary, the ATCS model does the following:

\begin{itemize}
    \item It allocates private resources to users belonging to a resource class $R\_Cl^k$;
    \item It shares unused public resources belonging to this class ($R\_Cl^k$) with users belonging to other resource classes independently of the resource class priority, and
    \item It allocates public available resources from other resource classes ($R\_Cl^z$ with $z\neq k$).
\end{itemize}

Figure \ref{fig:ATCS_Sharing} summarizes and illustrates the ATCS operation for private and public resource allocation and all resource classes.

ATCS users are mapped to resource classes $\sum_{k=1}^{n}R\_Cl^k$ (Figure \ref{fig:BAMParameters}). As such, for all nodes $n_j$ connected to node $n_i$ through links $l_{ij}$ with $i,j = 1,2, ..., n$, there are sets of users $\sum_{1=1}^{n}U^i$ mapped or grouped in resource classes $R\_Cl^k$.

In terms of the ATCS operation in a graph path with a set of nodes for each resource class $R\_Cl^k$, there is a User Allocated Resource matrix (UALR) with allocated resources requested by nodes $n_i$ to node $n_j$ through links $l_{ij}$ with $i,j = 1,2, ..., n$ (Equation \ref{eq:UALR})

\begin{equation} \label{eq:UALR}
\begin{split}
UALR=[\sum_{y=1}^{m}U^y\sum_{x=1}^{n}ALR^x]\forall R\_Cl^k, k=0,1,\dots,n.
\end{split}
\end{equation}

The ATCS model, in summary, allocates and shares resources among the users $U^y$ belonging to a resource class $R\_Cl^k$. The operation is such that unused resources for that class are shared until they are exhausted. This operational characteristic of the ATCS model is analytically modeled by equations  \ref{eq:ATCS_Operation} and \ref{eq:UALR} and illustrated by Figure \ref{fig:ATCS_Sharing}.

\section{ATCS Dynamics Use Case Simulation}\label{sec:ATCSUseCase}

The dynamics of the ATCS model operation (equations  \ref{eq:ATCS_Operation} and \ref{eq:UALR} and Figure \ref{fig:ATCS_Sharing}) may also be perceived by a case of use simulation scenario in which the resources available for a class are allocated until exhaustion.

The case of use simulation scenario is for the allocation of the resource bandwidth in a node, and the ATCS model configuration parameters are as follows:

\begin{itemize}
    \item ATCS model operating at node $n_j$ receiving requests from node $n_i$ (Figure \ref{fig:ATCS_Sharing});
    \item Total amount of resources allocated and shared $NRC_{ij}$ is 100 Mbps;
    \item Three resource classes ($R\_Cl^0$, $R\_Cl^1$, and $R\_Cl^2$);
    \item Resource class constraints are: $R\_CT^2_{ij}$ = 20 Mbps, $R\_CT^1_{ij}$ = 50 Mbps, and $R\_CT^0_{ij}$ = 30 Mbps;
\end{itemize}

In this didactic simulation scenario, users of different resource classes request resources sequentially from node $n_i$  to node $n_j$  as follows:

\begin{itemize}
    \item Request 1 - $R\_Cl^0$ - $R_{ij}R\_Cl^0N^j$ = 20 Mbps;
    \item Request 2 - $R\_Cl^1$ - $R_{ij}R\_Cl^1N^j$ = 30 Mbps;
    \item Request 3 - $R\_Cl^2$ - $R_{ij}R\_Cl^2N^j$ = 10 Mbps;
    \item Request 4 - $R\_Cl^0$ - $R_{ij}R\_Cl^0N^j$ = 30 Mbps;
    \item Request 5 - $R\_Cl^2$ - $R_{ij}R\_Cl^2N^j$ = 20 Mbps;
    \item Request 6 - $R\_Cl^1$ - $R_{ij}R\_Cl^1N^j$ = 20 Mbps;
    \item Request 7 - $R\_Cl^0$ - $R_{ij}R\_Cl^0N^j$ = 10 Mbps;
    \item Request 8 - $R\_Cl^2$ - $R_{ij}R\_Cl^2N^j$ = 10 Mbps;
    \item Request 9 - $R\_Cl^n$ - $R_{ij}R\_Cl^nN^j$ = any amount of resources;
\end{itemize}

As illustrated in Table \ref{Tab:LSPsEstabelecidas}, ATCS grants request 1, 2 and 3, allocating resources available at their respective resource classes (green highlight).

Request 4, belonging to resource class 0, extrapolates the class 0 capacity, and ATCS shares unused resources from resource class 1 (orange highlight). ATCS executes a similar operation with request 5 belonging to resource class 2 and shares unused resources from class 1 to grant the user request.

All following requests (6, 7, and 8)  can not be granted anymore by ATCS using private or unused resources from any class. In effect, private and unused resources are exhausted at this point.

This stage of the ATCS model execution is relevant concerning utilizing BAM models to contribute to reinforcement learning agent design and optimization.

There are two alternatives to consider at this execution point of the ATCS model in terms of the objectives and constraints of the resource allocation problem:

\begin{itemize}
    \item The resource allocation objective to be attained has a constraint not allowing allocated resources to be revoked from users (preemption or resource devolution); or
    \item The resource allocation objective allows allocated resources to be revoked from users.
\end{itemize}

\begin{figure*}[htbp]
    \centering
    \includegraphics[width=0.5\textwidth]{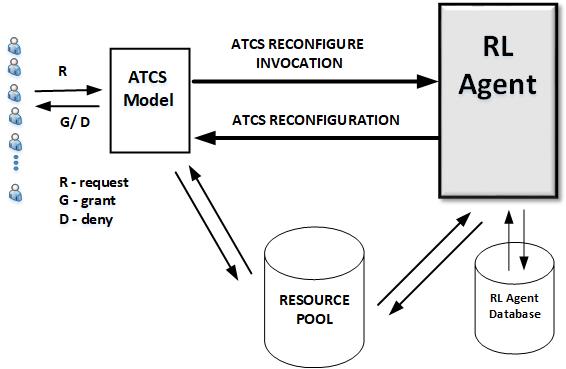}
    \caption{ATCS Offloading the RL Agent Resource Allocation Task.}
    \label{fig:ATCSandRLAgent}
\end{figure*}

If the first alternative prevails, the ATCS model must be managed, and additional resources are configured for the resource classes. This typically requires some management knowledge about the system's overall behavior. This task can be executed by an intelligent reinforcement learning agent or human manager.

If the second alternative prevails, the ATCS model preemption or devolution of resources is executed according to the resource class priority \cite{reale_alloctc-sharing_2011} \cite{martins_uma_2015}.

As an illustrative example of ATCS dynamic operation, request 6 from resource class 1 would imply preempting and returning resources used by request 4 and request 5 since they use resources that were shared because were unused but effectively belong to resource class 1. The following requests, 7 and 8, would be granted since resource classes 0 and 2 have unused resources at this phase of the resource allocation process.

As a final consideration, the arrival of request 9 from any class requesting any amount of resources would be denied since the maximum amount of resources for all classes is exhausted. There are no more resources to allocate or share whatsoever, and, as such, granting new resources to users imply in reconfiguring the ATCS parameters.

\section{ATCS Assisting the Design and Contributing to Optimizing the Reinforcement Learning Agent Operation}\label{sec:RLandBAMSupport}

A reinforcement learning agent design involves agent modeling and agent training. The RL agent training and performance depend significantly on the state-space model of the agent.

The issue addressed in this paper is whether the ATCS model contributes to the design and operational efficiency of an RL agent whose objective is to allocate resources among users.

We first argue that the ATCS model mimics an RL agent operation with similar allocating and sharing resources objectives. Consequently, the RL agent state-space is reduced since the BAM model does nearly all resource and sharing allocations until resources are exhausted. BAM operation, as such, facilitates the agent design and contributes to the optimization of its operation.

The ATCS model operation (including preemption and devolution) allocating and sharing resources is similar to an RL agent allocating resources when the following objectives are defined for both software:

\begin{itemize}
    \item ATCS and RL agent allocate resources to users based on resource priority;
    \item ATCS and RL agent allocate unused resources to all users; and
    \item ATCS and RL agent preempt allocated resources upon requests from high-priority resources or resource owners.
\end{itemize}

Equations \ref{eq:ATCS_Operation} and \ref{eq:UALR} of the analytical model demonstrate and validate the ATCS model mimicking the operation of an RL agent to allocate and share resources. The ATCS mimics the RL agent until:


\begin{itemize}
    \item Resources are exhausted for all ATCS resource classes; and
    \item Unused resources among classes are exhausted, and preemption and devolution are not allowed in the problem definition and requirements.
\end{itemize}

Based on this conclusion, the ATCS offloads and assists the RL agent computation of the resource allocation process, as proposed in Figure \ref{fig:ATCSandRLAgent}.



In the Figure \ref{fig:ATCSandRLAgent} setup, all resource-allocating computation is offloaded from the RL agent, and the agent acts on behalf of the ATCS model exclusively to manage its configuration. It is worth mentioning that the ATCS model management is a high-level task that requires knowledge acquisition and operation expertise in alignment with the purpose of an RL agent.

As such, the RL agent is invoked to manage the ATCS configuration parameters in a resource-exhausted condition. The RL agent is called because the ATCS model has no possibility whatsoever to allocate resources to users anymore when this condition is reached.

The RL agent design with ATCS assistance has the following benefits:

The RL agent allocating resources task offload inherently reduces the state-space model for the agent. That is so because it is invoked only to reconfigure ATCS parameters, and the agent no longer executes all computation and state analysis concerning user resource requests. The same reasoning applies to the agent modeling and training phases.

\subsection{The Computational Tradeoff between Agent Invocation and ATCS Algorithm Execution}

The reinforcement learning agent state-space reduction has a cost that is the deployment and processing overhead of the BAM algorithm (ATCS or any other).

In terms of computational complexity, the BAM software is inherently less complex than the RL agent that executes extensive state analysis and matrix processing consuming significant computational resources.


\section{Final Considerations}\label{sec:FinalConsiderations}

Reinforcement learning agents require advancements in terms of their design and efficiency. This paper proposes to use the ATCS BAM model integrated as part of an RL agent deployment to facilitate its design and contribute to the agent's efficiency.

The analytical model and the use case simulation presented demonstrate that an ATCS software offloads part of the computational tasks of a reinforcement learning agent aiming to allocate and share prioritized resources for users.

The RL agent task offloading is achieved by the ATCS model executing most of the allocation and sharing of resources. This approach results in the RL agent acting as a manager, adjusting the ATCS configuration parameters when resources are exhausted and need an intelligent management decision.

The proposed approach facilitates the agent design process, reduces the state-space of the agent, and, consequently, the training process involved.

Future work will include evaluating numerically the deployment and processing cost of the BAM algorithm against the cost reduction in modeling and training the RL agent in a reduced state-space.

\section*{Acknowledgment}

The authors thank the ANIMA Institute, Salvador University (UNIFACS), and the Federal Institute of Bahia (IFBA) for supporting this work.

\bibliographystyle{unsrt}
\bibliography{allbib}

\end{document}